\documentclass[dvipdfmx,12pt]{article}
\usepackage[utf8]{inputenc}
\usepackage[dvipdfmx]{graphicx}
\usepackage{amsmath}
\usepackage{amssymb}
\usepackage[size=a4paper]{bxpapersize}
\usepackage[top=30truemm,bottom=30truemm,left=25truemm,right=25truemm]{geometry}

\usepackage[sort]{natbib}

\usepackage{authblk}

\usepackage{hyperref}

\usepackage{makeidx}
\usepackage{url}
\usepackage{amsfonts}
\usepackage{bm,bbm}
\usepackage{natbib}
\usepackage{graphicx}
\usepackage{algorithm,algorithmic}

\newcommand{\tr}{\mathrm{Tr}}

\newcommand{\E}{\mathbb{E}}

\newcommand{\ve}{\varepsilon}

\DeclareMathOperator*{\argmax}{argmax}
\DeclareMathOperator*{\argmin}{argmin}

\newcommand{\cY}{\mathcal{Y}}
\newcommand{\cX}{\mathcal{X}}

\newcommand{\cD}{\mathcal{D}}

\newcommand{\cH}{\mathcal{H}}
\newcommand{\cL}{\mathcal{L}}

\newcommand{\cR}{\mathcal{R}}

\newcommand{\cV}{\mathcal{V}}

\newcommand{\bR}{\mathbb{R}}

\newcommand{\one}{\mathbbm{1}}

\usepackage{hyperref}

\title{Active Learning: Problem Settings and Recent Developments}
\author{Hideitsu Hino}
\affil{The Institute of Statistical Mathematics\\
10-3 Midori-cho, Tachikawa, Tokyo 190-8562, Japan \\
\tt{hino@ism.ac.jp}}

\date{}
\begin{document}
\maketitle

\abstract{
In supervised learning, acquiring labeled training data for a predictive model can be very costly, but acquiring a large amount of unlabeled data is often quite easy. Active learning is a method of obtaining predictive models with high precision at a limited cost through the adaptive selection of samples for labeling. This paper explains the basic problem settings of active learning and recent research trends. In particular, research on learning acquisition functions to select samples from the data for labeling, theoretical work on active learning algorithms, and stopping criteria for sequential data acquisition are highlighted. Application examples for material development and measurement are introduced.
}

\section{Introduction}
Supervised learning is a typical problem setting for machine learning that approximates the relationship between the input and output based on a given sets of input and output data. The accuracy of the approximation can be increased using more input and output data to build the model; however, obtaining the appropriate output for the input can be costly. A classic example is the crossbreeding of plants. The environmental conditions (e.g., average monthly temperature, type and amount of fertilizer used, watering conditions, weather) are the input, and the specific properties of the crops are the output. In this case, the controllable variables are related to the fertilizer and watering conditions, but it would take several months to years to perform experiments under various conditions and determine the optimal fertilizer composition and watering conditions. Methodologies to determine experimental protocols have been developed for obtaining the necessary information at a minimum cost, such as the experimental design method~\citep{fisher:1935,Hotelling1944,Lindley1956} and the sequential experimental design method~\citep{Johnson1961,Ford1980}. In the context of supervised machine learning, sequential experimental design is generally called {\it{active learning}}. Various methodologies have been proposed for the problem of sequentially selecting samples to improve the predictive performance based on a small amount of learning data available. In contrast to active learning, passive learning refers to when all data are given at once.

\begin{figure}[t] 
\centerline{\includegraphics[width=\columnwidth,clip]{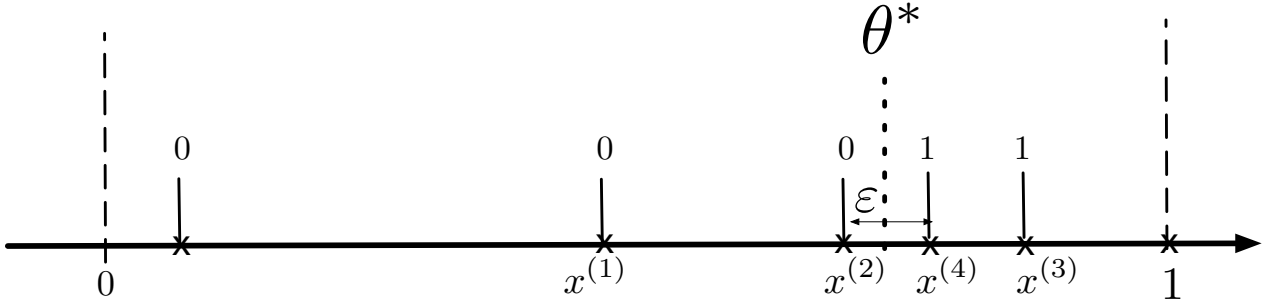}}
\vspace{-3mm}
\caption{Example of binary search for a split threshold of $x \in [0,1]$ that is completely separable. A two-class classifier $y = h_{\theta}(x) = \one(x \geq \theta)$ is efficiently learned.\label{fig:binsearch}}
\end{figure}
For active learning, proper selection of samples for training the predictor can reduce the probability of mistakenly predicting the response variable for an unknown explanatory variable (i.e., generalization error). Intuitively, the usefulness of active learning can be demonstrated by the following simple situation~\citep{Cohn1994}. If the variable $x$ is distributed in the interval $[0, 1]$, it is completely separable when $y = h_{\theta^{\ast}}(x) = \one (x \geq \theta^{\ast})$ with a certain threshold $\theta^{\ast} \in (0,1)$. According to standard Vapnik–Chervonenkis (VC) theory, a sample of $m = O(1/\ve)$ needs to be obtained from $\Pr(x, y)$ to yield a discriminator with an error of $\ve$ or less. In active learning, the predictor is trained with increasing the number of number of samples by sequentially selecting $x \in [0,1]$ and querying $y \in \{0,1\}$ for it; then, a hypothesis with a prediction error of $\ve$ or less can be efficiently obtained through a binary search. As shown in Figure~\ref{fig:binsearch}, $y$ is queried to obtain label $y = 0$ for the midpoint $x^{(1)}$ (i.e., half the interval $[0, 1]$). Then, label $y = 0$ is obtained for the midpoint $x^{(2)}$ between $x^{(1)}$ and $1$. Subsequently, label $y = 1$ is obtained for the midpoint $x^{(3)}$ between $x^{(2)}$ and $1$. Finally, label $y = 1$ is obtained for the midpoint $x^{(4)}$ between $x^{(2)}$ and $x^{(3)}$. The interval of $x^{(2)}$ and $x^{(4)}$ is $\ve$ or less. Thus, by making the midpoint the estimated value of $\theta^{\ast}$, the threshold $\theta^{\ast} \pm \ve$ can be searched. This binary search has a complexity of $O(\log (1/\ve))$ and is an example where active learning can be realized with exponentially fewer samples than passive learning but with the same accuracy\footnote{Generalized binary search~\citep{Nowak2008,Nowak2009}, which considers a situation with no observation noise (i.e., no uncertainty in the predicted label), uses the concept of adaptive submodularity, which has been shown to be nearly optimal~~\citep{Golovin2011}.}.

\begin{figure*}[t] 
\centerline{\includegraphics[width=\columnwidth,clip]{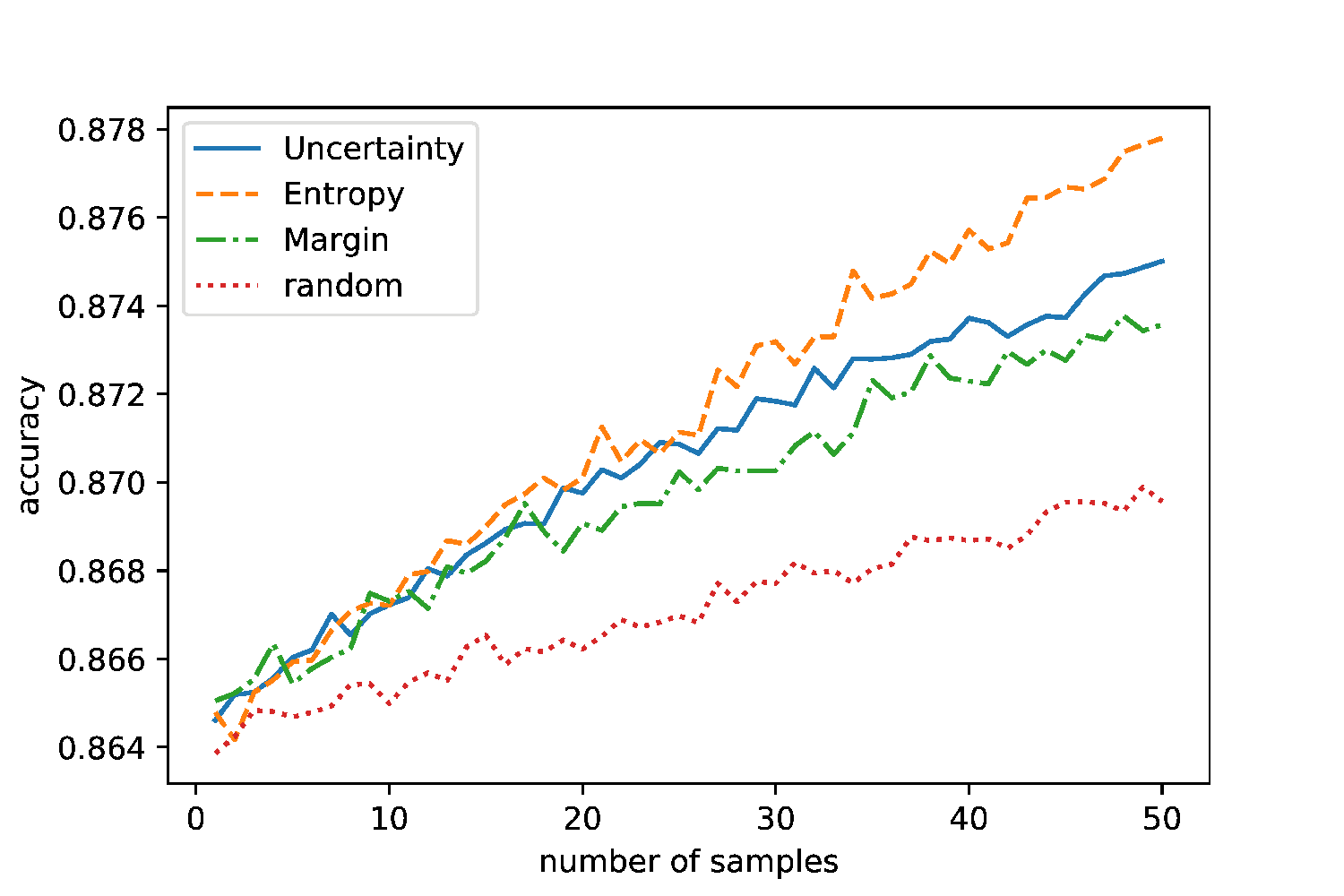}}
\vspace{-3mm}
\caption{Example of using active learning to predict MNIST data with random forest as the predictor. The horizontal axis represents the number of samples added to the initial 500 samples, and the vertical axis represents the prediction accuracy. The accuracy is observed to be improved with active learning compared with the accuracy of random sampling.\label{fig:exampleRF}}
\end{figure*}
This is a very simple one-dimensional two-class classification problem and is an ideal situation without noise. Although exponential acceleration is not possible in all cases, theoretical and experimental demonstrations have shown that active learning can achieve a high prediction accuracy with few samples. Figure~\ref{fig:exampleRF} compares some active learning methods with random sampling for the prediction of MNIST data~\citep{726791} with random forest~\citep{Breiman2001}. Specifically, the problem was to predict a $28 \times 28$ pixel handwritten number from $0 \sim 9$. The prediction model was trained with 500 sets of labeled data in advance. Fifty points of data were sequentially selected from the pooled data of $60,000-500=59,500$ points and added to the training data for the prediction model to learn, and the prediction accuracy was evaluated with a test dataset consisting of 10,000 points. {\tt{Uncertainty}}~\citep{Lewis1994}, {\tt{entropy}}, and {\tt{margin}}~\citep{Settles2010} represent the acquisition functions that were used to determine which sample was labeled next for active learning. Each learning curve shows the average results when the training dataset used for the initial prediction model was randomly changed 100 times. The graph shows that active learning improved the accuracy while using a smaller amount of data than random sampling.

This paper introduces the basic problem setting and concept of active learning, including recent research trends and application examples. Section 2 introduces terms and notations and explains the basic problem setting. Section 3 introduces typical criteria for selecting data that require labeling. Although various algorithms and acquisition functions have been proposed, it is not evident for which situations active learning is better than passive learning. In addition, some methods require solving an optimization problem that is difficult to execute in practice, even if the learning is efficient in principle. In such a case, an approximation method and guaranteeing the accuracy of the approximation are important. Accordingly, section 4 introduces results for theoretical guarantees of active learning. Section 5 presents research on the stopping criteria for active learning. Section 6 provides an example of applying active learning to measurement and material development. Finally, section 7 concludes the paper with future tasks and prospects.

We note that~\citet{Settles2010} presented a well-known survey of active learning. \citet{Hanneke2014} presented a survey focusing on the theoretical aspects when no assumption is made on the noise distribution, which is called {\it{agnostic}} active learning. \citet{Ramirez-Loaiza2017} presented an experimental comparison of various methods, and~\citet{Lowell2019} summarized the problems with practical application of active learning to natural language processing.

Active learning may be used in a broad sense to include Bayesian optimization. In this study, Bayesian optimization is considered as the search for optimal experimental settings (parameters or variables) from a small number of trials, while active learning is distinguished as improving the predictive model using a small amount of learning data.

\section{Problem Setting}
Let $X \in \cX$ be an explanatory variable and $Y \in \cY$ be a response variable. $\cY$ is a subset of $\bR$ for regression problems and is $\{+1,-1\}$ for discrimination problems. The realizations of the random variables $X, Y$ are denoted as $x$ and $y$, respectively. The function $h : \cX \to \cY$, which predicts the response variable from the explanatory variable, is called a hypothesis or predictor. The set of hypothesis spaces is represented by $\cH$. The mechanism for generating data, i.e., the true function $y = f(x)$, is called {\it{realizable}} when it is included in the assumed hypothesis space and non-realizable when it is not. Consider the random variables $(X, Y)$ with the joint distribution $D_{XY}$, and let the marginal distribution for $X$ of $D_{XY}$ be $D_X$. The data used for learning the hypothesis are $S_n = \{(x_1, y_1), (x_2, y_2), ..., (x_n, y_n)\} \in (\cX \times \cY)^n$. The loss function for evaluating the prediction error based on the hypothesis is represented by $\ell : \cH \times \cX \times \cY \to \bR_{+}$. In the case of a regression problem, the squared loss $\ell(h,x,y) = (y - h(x))^2$ is often used. In the case of a discrimination problem, the discriminant error or $0-1$ loss $\ell(h,x,y) = \one (h(x) \neq y)$ is often used. For active learning, discrepancies in discriminant results are easier to evaluate compared to discrepancies in regression results. In addition, theoretical analysis is easier with a version space that is described later, but the version space for regression problems does not have a clear definition. Thus, most studies on active learning have focused on discrimination problems. However, several active learning methods have been proposed for regression problems~\citep{KaySung1994,Krogh1995,Cohn1996,Hall2003,Burbidge2007,Castro2005}. In this paper, whether a situation is a discrimination or regression problem depends on the context and explicitly stated when necessary. 

For active learning, it is convenient to assume a subject (i.e., {\it{learner}}) who performs the learning for the predictor. The learner selects a sample $x$ for which the value of the corresponding explanatory variable $y$ is unknown by some criteria, thereby obtaining the value of $y$. The value of the explanatory variable is called a label for both discriminant and regression problems. The function that returns the value of the explanatory variable for $x$ is often called the {\it{oracle}}, but this implies that the label is always correct. In this paper, the term {\it{observation}} is used rather than oracle because, in practice, occasionally, only values that include labeling errors are obtained.

For distribution $D_{XY}$ on $\cX \times \cY$, the generalization error and empirical error are defined as
\[
\cL_{D_{XY}}(h) = \E_{D_{XY}} [ \ell(h,X,Y)], \quad
\cL_{S_{n}}(h) = \frac{1}{n} \sum_{i=1}^{n} \ell(h,x_i,y_i).
\]
Certain active learning methods for discrimination problems use the concept of {\it{version space}}:
\begin{equation}
    \cV(S_{n}) = \{
    h \in \cH | 
    h(x) = y, \; \forall (x,y) \in S_{n}
    \},
\end{equation}
which is a subset of the hypothesis set that is consistent with the training data obtained until that point.

In general, the prior distribution $p(h)$ is set for each hypothesis $h \in \cH$, and the hypothesis is stochastically selected. In Bayesian active learning, the posterior probability $p(h |S_n)$ of the hypothesis is learned with data $S_n$ to consider the inference given by the hypothesis sampled according to the posterior probability or the expected value given by the posterior probability of the predicted value~\citep{10.5555/143221}. A setting with the distribution included with respect to $h \in \cH$ is advantageous because the volume of the version space can be measured naturally.

The number of samples required for learning a hypothesis to obtain the desired prediction accuracy is called sample complexity in standard statistical learning theory. In the context of active learning, however, the required number of labels is the main concern and is sometimes called {\it{label complexity}}. In addition, there is some debate on how much unlabeled data is sufficient. In probably approximately correct (PAC) learning, terms such as $\log d, \log (1/\delta), \log \log (1/\ve)$ often appear in relation to parameters such as the VC dimension $d$, prediction error $\ve$, and confidence $\delta$. However, in this paper, such terms are omitted in favor of order notation, such as $O(n)$ and $O(\log n)$.
In statistics, most sequential designs assume a setting where observation points can be freely selected according to some standard; this is called membership query synthesis in the context of active learning~\citep{Angluin1988}. This is acceptable when the corresponding explanatory variable value can be reliably obtained. However, for example, this is difficult to apply to problems involving optical character recognition (e.g., appropriate labeling of characters freely composed by the learner)\footnote{In character recognition research, the approach of synthesizing and learning character images on a computer is widely taken, and in this case, the label of the synthesized image is known. On the other hand, in active learning, an unlabeled synthetic image that is judged to belong to the boundary region of the character type, which is difficult to label, is generated, and labeling is required for that image.}. Thus, two types of active learning are usually considered for machine learning, as detailed in the subsequent subsections.

\subsection{Stream-based active learning}
In stream-based active learning, data are sequentially presented from the data  generating distribution $D_{X}$ to the learner, who decides whether to request label $y$ for the presented data $x$ based on some criterion. If a label is not requested, then the data are discarded. Labeled data are used to train the predictor. There may be an option where the labeled data are not used to train the predictor. \citet{Atlas1990} presented a typical discussion of stream-based active learning in the field of machine learning. For a discrimination problem, if there is a region in the version space where the predicted label differs from the presented sample, it is possible to query samples that are difficult to predict with the current hypothesis by labeling the sample with a high degree of uncertainty. An exact calculation of the version space and its sub-regions is difficult, hence approximation methods have been proposed. For example, ~\citet{Atlas1990} proposed a method of constructing a filter that uses a neural network to extract samples that do not match the predictions included in the version space.

\subsection{Pool-based active learning}
In pool-based active learning, a small number of pre-labeled datasets and rich unlabeled datasets (i.e., pooled data) are provided, and a predictor trained with a small dataset is used to select samples from the pooled data for labeling. The data selected for labeling are added to the pre-labeled data available, and the predictor is trained again. This setting represents a typical problem where data collection is easy but the annotation cost is high.
Although numerous active learning methods assume that only one query is issued at a time, retraining the model every time a sample is added is inefficient. The time required for one learning session may not be neglected, or the addition of a single sample may not make a meaningful change in the model, especially for deep learning models. \citet{Cohn1994} suggested batch active learning because of the high complexity of version space management. In situations where multiple experiments are parallelly conducted, batch active learning is effective because multiple samples from the pooled data can be selected and labeled simultaneously. If a batch of size $k$ can be labeled at once, a simple strategy may be to select the top $k$ samples by repeatedly selecting a single sample $k$ times; however, a concern is that only $k$ similar samples may be selected. For batch active learning, it is crucial for the selection criteria to consider sample diversity as well as the amount of information that each sample contains with respect to the model. Various methods have been proposed, such as a method based on the notion of submodularity~\citep{Hoi2006} as described below, and formulating the batch selection problem as a non-convex integer program for empirical risk minimization~\citep{Wang2015}.
Problem settings can be conveniently classified as stream-based, pool-based, or arbitrary design; often, a method that assumes a certain setting can still be applied to another setting with slight modifications.

\section{Acquisition Function}
Perhaps the most important component of active learning is the acquisition function, which determines whether a sample requires labeling. Several acquisition functions aim to quantify the difficulty of label prediction in some way and actively incorporate difficult-to-predict samples into learning. Numerous early studies on active learning designed the acquisition function through heuristics and presented theoretical analyses. In addition, several methods have recently proposed using reinforcement learning or transfer learning to obtain acquisition functions according to the environment and data.

\subsection{Design of acquisition function}
Most active learning algorithms use an acquisition function that selects a more informative sample or representative sample. The most intuitive approach is to choose the most uncertain data for the current hypothesis. This is called uncertainty sampling and is based on the expectation that if the current prediction model labels the most uncertain data, then the model uncertainty will be reduced the most~\citep{Lewis1994a,Yang2015}. Various measures of uncertainty are available depending on the problem and model. For logistic regression and multinomial logistic regression models, where the probability value naturally accompanies the prediction, all data that are query candidates are predicted, and the sample with the prediction probability closest to $0.5$ (for two-class classification) or with high entropy in the predicted distribution (entropy sampling) can be selected. Alternatively, a sample can be selected that maximizes the difference in probabilities of the labels with the highest and second-highest probabilities (margin sampling). Another approach is to choose the sample with the smallest maximum prediction probability (i.e., least confident sampling). Probabilistic models are a standard means of measuring the prediction uncertainty. Nevertheless, there are also methods that use the reciprocal of the support vector machine (SVM) margin~\citep{Tong2002}, where the hypothesis can be efficiently narrowed down by dividing the version space into two parts with approximately equal volumes for each query.
To evaluate uncertainty, a standard approach is to learn a hypothesis that provides a probabilistic output or to use an ensemble of several hypotheses. However, such prediction models are not always suitable for learning. \citet{Lewis1994} achieved a low error rate with significantly fewer samples than random sampling by performing active learning using a different model from the one used for prediction. \citet{Houlsby2011} proposed a method of issuing queries under a Bayesian setting where the entropy of the predicted values is high and the expected entropy of predicted values for the parameter posterior distribution of the prediction model is low (i.e., low uncertainty of predicted values with individual parameter settings). They called their method Bayesian active learning by disagreement (BALD). This method was extended by \citet{Kirsch2019} for batch active deep learning, (BatchBALD), and it is currently considered one of the best-performing methods available.

Given the premise of ensemble learning, the uncertainty of a prediction can be defined as the sample with the most split votes. \cite{Seung1992} studied the most basic form of query by committee (QBC), where an even number of hypotheses is learned, and samples are queried such that the prediction results of two-class labels fall in two halves. They proposed a method of sampling a hypothesis from the version space for unlabeled samples and issuing a query when the prediction results do not match. Intuitively, a sample is difficult to predict when multiple elements of the version space corresponding to the hypothesis are extracted and do not match the prediction results. Moreover, the learning efficiency can be improved if the corresponding label is obtained. \citet{Seung1992} used a method of statistical physics to show that the generalization error decreases exponentially with the number of queries in the limit of an infinite number of hypotheses that are learned simultaneously. \citet{Freund1992} extended the analytical approach of \citet{Seung1992} to a broader class of discrimination problems. They showed that, under certain conditions, the generalization error decreases exponentially with the increase of labeling. Because QBC involves sampling hypotheses from the version space, rigorous implementation is difficult. \citet{Gilad-Bachrach2005} proposed a method for efficiently sampling hypotheses by projecting a version space onto a low-dimensional space. Various acquisition functions are available~\citep{Settles2010}, such as selecting the query that changes the model majorly~\citep{Cai2013a} or minimizes the approximation of the expected error~\citep{Guo2007}.

The acquisition functions introduced so far have focused on selecting samples with a large amount of information with regard to model uncertainty and change. Because this approach does not consider the distribution of unlabeled data, there is a risk of issuing queries that are considerably affected by the discriminant surface obtained from the current hypothesis. 
It is pointed out by~\citet{Eisenberg1990OnTS} that it is only a limited situation that active learning improves sample complexity in terms of PAC learning theory. One explanation for this is that random sampling contains information on hypotheses (i.e., correct mapping of explanatory variables to response variables) and on the distribution of explanatory variables, which is lost with active learning~\citep{Freund1992}. It is reasonable to require active learning strategy to reflect the distribution of explanatory variables (e.g., actively sampling points with high density as representative), and methods have been proposed that are based on the idea of annotating representative samples. \citet{Nguyen2004} selected representative samples by clustering pooled data in advance, and \citet{Settles2008} implemented the same by weighting data according to the $k$-neighbor distance. However, selecting a representative sample requires a large number of queries compared to an approach that focuses on the amount of information, such as uncertainty. Some studies have considered combining the two approaches to construct efficient and effective acquisition functions~\citep{Xu2003,Donmez2007,Huang2014}). For example, \citet{Huang2014} considered a sample with a small margin for a two-class discrimination problem to have high uncertainty; they selected a representative and informative query by solving the minimax optimization problem of minimizing the risk for all possible labels of pooled data, while maximizing the risk of labeling candidate samples.

\subsection{Learning the acquisition function}
The success or failure of active learning with the acquisition functions introduced in the previous subsection depends on the prediction model, data distribution, and compatibility of the acquisition function to them. In addition to the selection of the prediction model, another problem is the general difficulty of selecting an appropriate acquisition function. Accordingly, an approach called meta-active learning has recently been proposed, where an acquisition function is learned from data~\citep{Konyushkova2017}. A promising approach is to formulate active learning in the framework of reinforcement learning and express the acquisition function as a policy to be learned by reinforcement learning~\citep{Ebert2012}.

For the reinforcement learning of a typical acquisition function, some methods approximate the Q-function with a deep Q-network (DQN)~\citep{Mnih2015}. \citet{Fang2017} regarded stream-based active learning as a Markov decision process and proposed learning the optimal policy by setting the parameter $\theta$ of the prediction model and represented {\it{state}} as the unlabeled sample $x$ and {\it{action}} as whether or not labeling is required. \citet{Woodward2017} used one-shot learning with neural Turing machines to design states, behaviors, strategies, and rewards; in addition, they used reinforcement learning to design a function that determines if a label needs to be requested for the presented sample for stream-based active learning. By using deep reinforcement learning with long short-term memory (LSTM) as the Q-function to determine the value of an action in a certain state, it is possible to judge whether a sample is unknown and whether a label should be requested. Specifically, the {\it{state}} (or {\it{observation}}) is a pair of labels for the input and the previous observation. It is used to determine whether to request a label for the currently given input, depending on the confidence level of the prediction for the input and if a label was requested for the previous input. The {\it{action}} is to request a label for the presented sample or predict the label itself. If the prediction of the learner of the label is correct, the reward is zero; if it is incorrect, the reward is negative. If the learner requests a label, a slightly negative reward is obtained. This allows a policy to be learned so that a label is not required if there is confidence in the prediction. Furthermore, \citet{Bachman2017} applied this method to reinforcement learning of the acquisition function based on matching networks for pool-based active learning.

The problem of learning a policy for selecting the optimal query in an ever-changing environment is closely related to the bandit problem. \citet{Baram2004,Hsu2015} considered the design problem of the acquisition function for active learning as a multi-armed bandit problem. They proposed a method to select the optimal strategy as multiple acquisition functions or a linear combination of such functions. \citet{Chu2017} further proposed a method of using a strategy learned as a multi-armed bandit problem in one domain for active learning in a different domain. \citet{Wassermann2019} proposed reinforcement learning of a filter for stream-based active learning based on reinforcement learning. Deep reinforcement learning has also been used to change the acquisition function dynamically to follow changes in the input distribution. \citet{Haussmann2019} used a Bayesian deep learner as a model for evaluating uncertainty and defined states with its predicted distribution. In addition, they proposed a method where another Bayesian deep learning model is learned separately from the output prediction, and the current predictor state and the acquisition function appropriate for the data are used as feature quantities together with the predicted output distribution and data.

Large-scale models for active learning are not expected to change significantly if only a single data point is added to the training data~\citep{Sener2018}. In addition, because of the high learning cost of the model, batch active learning is considered as a suitable approach. \citet{Ravi2018} used meta-learning for batch active learning of acquisition functions. \citep{Sener2018} empirically showed that the classical active learning of acquisition functions based on heuristics is ineffective for a convolutional neural network (CNN). They formulated batch selection as a core-set selection problem, which is defined as selecting a group of data points for learning such that the predictor trained using the selected subset is not significantly different from the case where all data points are used for learning. To solve the core-set selection problem with no label information, they derived an upper bound for the difference between the average loss for a subset and the average loss for all data. Subsequently, they developed an active learning algorithm that minimizes the upper bound.
Attempts to apply active learning to large-scale models such as deep learning have been actively studied in recent years. \citet{Sinha2019} designs an acquisition function using a variational autoencoder (VAE)~\citep{Kingma2014} and an adversarial network~\citep{NIPS2014_5ca3e9b1}. Specifically, the VAE is trained to match the distribution of labeled and unlabeled data, and adversarial networks are labeled against the latent space of the VAE. It is learned to determine whether or not it corresponds to the data already given. By using the discrimination network of the adversarial network as the acquisition function, the learning of the acquisition function that performs query selection independent of the task is realized.

\section{Theoretical Guarantee}
Active learning is a type of feedback system. Because samples are added through the acquisition function, training samples cannot be expected to follow a independent and identical distribution, which is the assumption of standard learning theory and statistical analysis. The generalization error can be reduced exponentially by reducing the version space, as in QBC for a noise-free discrimination problem~\citep{Freund1997}. The generalization error for practical algorithms is still being actively researched. In several cases, the acquisition function is defined so that some criterion is maximized or minimized. However, this optimization is often an NP-hard problem that is difficult to solve. This section introduces an approach that uses submodularity to provide a theoretical guarantee when an optimization problem associated with the evaluation of an acquisition function is approximated by a greedy algorithm. Also, researches on evaluation of label complexity are briefly summarized.

\subsection{Utilization of submodularity}
The concept of submodularity with respect to a set function~\citep{fujishige1991submodular} is useful for characterizing the effect of gradually adding training data, such as in active learning. Suppose there is a support set $V$ and power set function $2^{V}$, and a set function $f: 2^{V} \to \bR$. Here, the set function $f$ is considered as an index that expresses the value when certain elements are selected from $V$. Although there are some equivalent expressions, we adopt the definition of submodularity which is suitable for the purpose of this paper.
For any $S \subseteq V,\; v \in V$, $f(v | S ) = f( \{v\} \cup S) - f(S)$ is called the increase by $v$ relative to a set $S$. For an arbitrary $A \subseteq B \subset V$ and $v \in V \backslash B$, when $f(A \cup \{v\} ) - f(A) \geq  f(B \cup \{v\}) - f(B)$ or equivalently $f(v|A) \geq f(v|B)$ holds true, the set function $f$ is called submodular. Intuitively, the value of the newly added element $v$ decreases as the existing element set increases in size. When the function $f$ further satisfies $f(v|A) \geq 0$, $f$ is called monotonous. Consider the problem of maximizing $f(A_{k})$ by sequentially extracting $k$ elements from $V$ to construct a subset $A_k$. For a monotonic submodular function with $f(\emptyset)=0$, the simple policy $A_{i+1} = A_{i} \cup \{ \argmax_{e\in V \backslash A_{i}} f(A_{i} \cup \{e\})\}$, where a greedy element is added solely based on the value of $f$, guarantees an approximation ratio of $1-1/e$~\citep{Nemhauser1978}. In other words, $f(A_{k}) \geq (1-1/e) \max_{|A|\leq k} f(A)$.
Because the value obtained by acquiring new labels gradually decreases in pool-based active learning, an algorithm using submodularity is suitable for batch active learning. If the batch size $k$ is fixed, the problem of selecting $k$ samples to maximize the entropy of the predicted distribution $p(y|x)$ is guaranteed to have an approximation rate of $1-1/e$ through selection by the greedy algorithm based on the submodularity of entropy. \citet{Hoi2006} considered a two-class discrimination problem using a model with the parameter $\bm{w}$ as a hypothesis, and quantified the effect of batch data on the model by using a Fisher information matrix
\begin{equation}
I_{p}(\bm{w}) = - \E_{p} \sum_{y = \pm 1} p(y|x) \frac{\partial^2}{\partial \bm{w}^2} \log p(y|x) dx
\end{equation}
for the distribution $p(x)$ of the sample $x$. Specifically, they aimed to select a batch that was as close as possible to the information that all samples had, by minimizing $\tr [I_q(\bm{w})^{-1} I_p(\bm{w})]$, where $p(x)$ and $q(x)$ represent the pooled data distribution and selected batch distribution, respectively. This criterion is equivalent to asymptotically minimizing the expected squared error and has long been used as an acquisition function for active learning~\citep{Cohn1996,374276,FUKUMIZU1996}. \citet{Hoi2006} showed that an approximation of $\tr [I_q(\bm{w})^{-1} I_p(\bm{w})]$ for a logistic regression model is a monotonic submodular function and that the selection of batch data by the greedy algorithm is a suitable approximation of the optimal batch. Thus, by designing an acquisition function with submodularity for batch active learning, batch data can be efficiently acquired in each round by the greedy algorithm at a guaranteed approximation rate. However, the concept of submodularity is insufficient for describing the selection of the optimal sample according to data acquired so far and the hypothesis. Some research has characterized the sample acquisition policy by generalizing the notion of submodularity, which is called {\it{adaptive submodularity}}. \citet{DBLP:conf/colt/GolovinK10} showed that reducing the version space can be regarded as a covering problem for a set of unsuitable hypotheses, which has adaptive submodularity. They utilized this fact to analyze the label complexity of a generalized binary search algorithm when there is no noise in the observation. When there is noise, \cite{Golovin2010} considered the problem of seeking a query strategy that minimizes the cumulative cost of labeling. By selecting samples in a finite hypothesis space until the size of the version space becomes unity, a unique hypothesis can be identified. They proposed an efficient approximation algorithm based on adaptive submodularity. Similarly, the adaptive submodularity of version space reduction has been used to derive approximation rates for batch active learning with a greedy algorithm and pool-based active learning using several criteria, such as entropy maximization~\citep{Chen2013,Nguyen2013}. The concept of adaptive submodularity is originally for pool-based active learning, but \citet{Fujii2016} extended it as a policy adaptive submodularity, and derived a mini-batch active learning algorithm and a stream-based active learning algorithm with guarantees of the approximation rate.

\subsection{Learning theory of active learning}
Determining when active learning reduces the generalization error at a faster rate than passive learning is an important question and has been the subject of various studies. Although active learning can be applied to the binary search introduced in section 1, it does not work well if there is even a small amount of observation noise. \citet{Eisenberg1990OnTS} stated that queries often have little effect on label complexity in the framework of PAC learning.
Nevertheless, active learning has been clearly demonstrated to improve label complexity in noisy observation cases. Regarding problems where active learning is no better than passive learning,~\citep{Balcan2008} indicated that the sample complexity required to train a predictor with a generalization error of $\ve$ or less differs from the sample complexity required to guarantee that the resulting predictor has a generalization error of $\ve$ or less.  Subsequently, they analyzed situations where active learning can effectively reduce label complexity.

Representative studies on the label complexity of active learning algorithms are found in \citep{Dasgupta2005,Dasgupta2005b}. \cite{Dasgupta2005} presented a simple example: for a realizable and one-dimensional case, learning by $O(\log (1/\ve))$ through a binary search is possible; for a case with two dimensions or more, $O(1/\ve)$ is unavoidable in the worst case (i.e., the correct solution is an unbalanced hypothesis). The study shows that $O(d \log (1/\ve))$ is possible as an average for an appropriate distribution in the hypothesis space, where $d$ is the VC dimension of the hypothesis space. For the problem of two-class discrimination without noise, \citet{Dasgupta2005b}) introduced the following three quantities: desired accuracy $\ve$, the degree to which the size of the version space is reduced by a single query $\rho$, and the quantity $\tau$ to characterize the amount of unlabeled data (i.e., pooled data). Then, defining the notion of  $(\rho,\ve,\tau)$ separability of the hypothesis space $\cH$, active learning algorithm was characterized. By using $\rho$, \citet{Dasgupta2005b} introduced an example that the sample complexity of active learning is between $O(1/\ve)$ and $O(\log (1/\ve))$, and for the realizable hypothesis space, $O(\log (1/\ve))$ can be achieved depending on the size of the data distribution and pooled data.

In online active learning, when the problem is linearly separable and the data is in the $d$ dimensional unit sphere, \citet{Freund1997} showed that lable complexity for solving a realizable problem if of order $O(d \log (1/\ve))$, and the number of discarded samples without labeling is of order $O((d/\ve) \log (1/\ve))$. In QBC, it is generally difficult to sample hypotheses from the version space. \citet{Cesa-Bianchi2003,Dasgupta2009} showed that $O(d \log (1/\ve))$ is realizable with a simpler algorithm than QBC for data distributed on a unit sphere.

Much of the discussion about sample complexity in active learning has been on reducing version space, and the hypothesis space has been assumed to contain a true hypothesis (otherwise the version space can be an empty set). \citet{Kaariainen2010} discussed the label complexity of active learning in unrealizable situations and showed that this typically did not lead to exponential improvement compared to passive learning. In ordinary statistical learning theory, faster convergence can be obtained by imposing conditions on the observation noise rather than using active learning. Even with active learning, imposing restrictions on conditional distributions, such as the margin conditions, may result in exponentially faster convergence than with passive learning. On the other hand, agnostic active learning can be used to deal with situations where observation noise is not assumed~\citep{Balcan2006,Hanneke2007,Dasgupta2008}. \citet{Balcan2006} proposed an $A^2$ algorithm that operates under arbitrary noise conditions and achieved a sample complexity of $O(\eta^{2}/\ve^2)$, where $\eta$ is the generalization error due to the optimal hypothesis in the assumed hypothesis space. The learner must approach the optimal hypothesis in the hypothesis space as close as $\ve$. \citet{Hanneke2007} introduced the discrepancy coefficient $\theta$, which is determined by the hypothesis space and data distribution and showed that the sample complexity of the $A^2$ algorithm by \citep{Balcan2006} is of order $O(\theta^2 \eta^2/\ve^2)$. This was further improved to $O(\theta \eta^2/\ve^2)$ by~\citet{Dasgupta2008}. Notably, noise is not the only reason for a case to be non-realizable. Research on sample complexity has also considered noise models, such as the margin condition~\citep{Castro2008}.

Not many studies on label complexity have focused on computational efficiency. \citet{10.1145/3006384} discussed a methodology that guarantees the acquisition of the error $\ve$ in polynomial time with bounded noise and agnostic settings. The theoretical analysis of active learning is detailed in a tutorial by Hanneke and Nowak{\footnote{\url{https://nowak.ece.wisc.edu/ActiveML.html}}}.

\section{Stopping problem}
For sequential algorithms, the timing to stop learning is generally an important issue. For active learning, the optimal stopping criterion depends on the tradeoff between the labeling cost and the gain of improving the prediction accuracy. If the limit of the measurement technology (noise level) is known or the theoretically achievable error can be estimated, learning may be stopped when the prediction accuracy comes close to that level. However, even if the labeling cost is not clearly determined or the target accuracy is not clearly defined, there may be situations where active learning is desirable to reduce learning costs. In the case of a fixed budget for labeling (i.e., fixed number of times), requesting the full budget of labels may seem the best approach. However, if sufficient learning data have already been obtained within the budget, a smaller budget may be sufficient for similar situations in the future. In addition, determining the room for improvement of the prediction accuracy when the budget is exhausted is important. A simple approach to determine whether the maximum generalization error has reached its peak is to evaluate the prediction performance with cross-validation or holdout data. However, active learning is generally assumed to be for situations with high labeling costs, and obtaining holdout and validation data separately cannot be expected. In other words, determining when to stop active learning requires self-contained statistics.
The optimal stopping problem has been considered in the field of OR~\citep{Gilbert1966,Cisse2012}. In addition, early stopping~\citep{Prechelt2012} is widely used with regularization methods, especially for the learning of complex models, such as deep learning. For kernel ridge regression, early stopping has an implicit regularization effect~\citep{Yao2007,Rosasco2015}, and its theoretical properties have been considered for nonparametric regression~\citep{Raskutti2014}. Some heuristics have been proposed in the context of Bayesian optimization~\citep{Desautels2014,Lorenz2015}, but few have a solid theoretical background~~\citep{Dai2019}. Particularly for active learning, the {\it{less is more}} phenomenon~\citep{Schohn2000} has shown that a model trained with a small amount of data outperforms a model trained with all the data. This is the motivation for active learning and shows the significance of determining the optimal time to stop learning.
Methods for determining the stopping criterion for active learning include using the entropy of the prediction result for the pooled data, changes in the prediction result, and certainty of the prediction~\citep{Zhu2008}; or considering the stability of prediction results by multiple predictors~\citep{Bloodgood2013,Altschuler2019}. Other methods include stopping when the sample inside the SVM margin disappears from the pooled data~\citep{Schohn2000,Vlachos2008} and checking the convergence of a similar quantities~\citep{Laws2008,Krause2007}, but their theoretical bases have not been fully verified.

\citet{Krause2007} focused on batch-mode Bayesian active learning with Gaussian process regression; they used submodularity to derive a relation among mutual information for the predicted distributions when samples are actively selected or randomly selected. Although their method has been suggested as a stopping criterion for active learning, it significantly relies on the prediction model being a Gaussian process regression, and it also requires appropriate discretization of the domain of an explanatory variable.
The $A^2$ algorithm of \citet{Balcan2006} is formulated as the following procedure according to~\cite{Hanneke2014}. For the $t$-th iteration of the algorithm, $2t$ unlabeled samples are selected and set as $S$. A query is issued, and a label is provided to the intersection $Q=DIS(\cH) \cap S$ of the disagreement subspace
\begin{align*}
DIS(\cH) = 
\{
x \in \cX |
\exists h, h^{\prime} \in \cH ,\; 
h(x) \neq h^{\prime}(x)
\}
\end{align*}
for the hypothesis space and $S$.
By using the labeled dataset $Q$, the empirical loss $\cL_Q(h)$ is minimized to select the hypothesis $h \in \cH$. Namely, $\hat{h} = \argmin_{h \in \cH} \cL_{Q}(h)$. Finally, a hypothesis $h$ is removed from the hypothesis space if it satisfies
\[
\cL_{Q}(h) - \cL_{Q}(\hat{h}) >
\sqrt{ \hat{P}_{Q}(h \neq \hat{h}) \frac{d}{|Q|}}.
\]
Here, $\hat{P}_{Q}(h \neq \hat{h})$ is the estimated probability using $Q$ that $h$ is not equal to the hypothesis $\hat{h}$, and $d$ is the VC dimension of the hypothesis space. By repeating this process and reducing the hypothesis space, a highly accurate hypothesis remains. Using this algorithm, it is natural to stop learning if $\max_{h \in \cH} \sqrt{\hat{P}_{Q}(h\neq \hat{h})\frac{d}{|Q|}} < \epsilon$ with an appropriate threshold $\epsilon>0$.

In a more general approach, \citet{DBLP:conf/aistats/IshibashiH20} tackled the problem of stopping active learning from the perspective of PAC Bayesian learning~\citep{DBLP:journals/ml/McAllester99} when the posterior distribution is obtained. The range of the loss function $\ell$ is assumed to be $[a,b], \; -\infty < a < b < \infty$. In the PAC Bayesian learning framework, hypothesis $h$ also follows the distribution of function values, and a prior distribution $p(h)$ is introduced. Let $p(h | S_{n})$ be the posterior distribution of $h\in \cH$. If the negative log-likelihood $\ell$ is the loss function, then $\ell(h,x,y) = - \log p(y|h,x)$ and $p(y|h,x) = e^{-\ell(h,x,y)}$. The posterior distribution of $h$ obtained using the training data $S_n$ is
\begin{align*}
    p(h | S_{n}) =&
    \frac{p(\bm{y}_{n}|h) p(h)}{p(\bm{y}_{n})} 
    =
    \frac{e^{-n \cL_{S_{n}}(h)} p(h) }{p(\bm{y}_{n})}
\end{align*}
where the values of the response variable included in $S_n$ are arranged as the vector $\bm{y}_{n}$. The predictive distribution for $y_{\ast}$ corresponding to the new input point $x_{\ast}$ can be represented as
\begin{align*}
    p(y_{\ast}|S_{n},x_{\ast}) =&
    \int p(y_{\ast} |h,x_{\ast}) p( h| S_{n}) d h.
\end{align*}
The posterior distributions of the hypotheses obtained by learning with $S_{n}$ and $S_{n+1} = S_{n} \cup \{(x_{n+1}, y_{n+1})\}$ are $p(h|S_{n})$ and $p(h|S_{n+1})$, respectively. The difference with the expected values for the posterior distribution of the expected risk is defined as
\begin{equation}
\cR ( p(h|S_{n}) , p(h|S_{n+1})) 
=
\E_{p(h|S_{n})} [ \cL_{D}(h)] - 
\E_{p(h|S_{n+1})} [\cL_{D}(h)],
\end{equation}
where $\cR$ represents the amount of reduction in the expected generalization error. When $p(h|S_{0}) = p(h)$, then
\begin{align*}
    \cR(p(h) | p(h|S_{n})) =&
    \sum_{i=1}^{n} \cR(p(h|S_{i-1}), p(h|S_{i})) \\
    =&
    \E_{p(h)} [\cL_{D}(h)] - 
    \E_{p(h|S_{n})} [\cL_{D}(h)] = 
    const. - \E_{p(h|S_{n})} [ \cL_{D}(h)]
\end{align*}
and the convergence of $\cR$ and $\E_{p(h|S_{n}) }[\cL_{D}(h)]$ is equivalent. If $p(h|S_{n})$ and $p(h|S_{n+1})$ are the posterior distributions of $h\in \cH$ given $S_{n}$ and $S_{n+1}$, then the following holds true:
\begin{equation}
\label{eq:UB}
    \cR(p(h|S_{n}), p(h|S_{n+1})) 
    \leq 
    D_{KL}[p(h|S_{n}) || p(h|S_{n+1})] + C.
\end{equation}
Here, $D_{KL}$ is the Kullback–Leibler (KL) divergence, and $C$ is a quantity that depends on the range $[a, b]$ that $h \in \cH$ can take. The KL divergence on the right hand side is a quantity that can be calculated for a concrete predictor (e.g., Gaussian process) under certain conditions. \citet{DBLP:conf/aistats/IshibashiH20} proposed a method to determine the optimal time to stop active learning by sequentially calculating the upper bound of $\cR$ and combining convergence tests for pool-based active learning when Gaussian process regression is the predictor and the acquisition function is the predictive variance.
Note that the upper bound of Eq.~\eqref{eq:UB} does not become less than the constant term because KL divergence is nonnegative. Therefore, when the constant term becomes large, the upper bound does not become tight. Another upper bound can be derived that is tight even if the constant term is large as long as the KL divergence is sufficiently small. Let $p(h|S)$ and $p(h|S^{\prime})$ be posterior distributions for the hypothesis $h \in \cH$ given $S$ and $S^{\prime}$. At this time, the following inequality holds for any $\lambda>0$ and any measurable function $\mathcal{L}_{\cD}(f)$ and $v\geq\mathbb{E}[\mathcal{L}^{2}_{\cD}(f)]$:
\begin{equation}
    \cR(p(h |S),p(h |S')) \leq \frac{v}{b-a}(e^{W(u)+1}-1).
    \label{eq_upper_bound_lambert}
\end{equation}
Here, $W(\cdot)$ is the Lambert $W$ function, and $u = \frac{(b-a)^2 D_{KL}[p(h |S)||p(h |S')]-v}{ve}$. For the bound of Eq.~\eqref{eq_upper_bound_lambert}, $W\left(-1/e\right)=-1$ when the KL divergence is zero. This guarantees that the upper bound also approaches zero as the KL divergence approaches zero, making it a tight bound.

\section{Application to measurement and material development}

Since the establishment of the US Materials Genome Initiative, research has been actively done to improve the efficiency, acceleration, and sophistication of materials and measurements through machine learning. Prototyping high-performance materials and measurements in synchrotron radiation facilities is often time-consuming and costly. Therefore, active learning is expected to substantially improve efficiency~\citep{Lookman2019,DelRosario2020,Tian2020}. This section introduces two example applications of active learning to material searching and measurement.

\subsection{Efficient phase diagram creation}
A phase diagram or state diagram represents the relationship between the state of a material or system and thermodynamic parameters. For example, $H_2O$ takes one of three states depending on the temperature and pressure: solid, gas, and liquid. For metallic materials, the microstructure can be austenite, ferrite, or pearlite depending on the mixture ratio of raw materials, processing temperature, and annealing pattern. Understanding the relationship between these different states and the thermodynamic parameters is necessary to understand the material properties and develop materials with desirable properties. An important first step for material development is running trials and measuring materials while varying different parameters to create phase diagrams. Comprehensive prototyping and the measurement of multiple parameters are often bottlenecks for the development process. \citet{Terayama2019} proposed using active learning to create phase diagrams efficiently. In their method, the range of the target parameter is divided into an appropriate grid resolution, and the state is measured at a few points on the grid. In terms of the notation of active learning, the grid is $\cX$, and the parameter configuration represented by each point corresponds to $x \in \cX$. The state at each point $x$ corresponds to $y$. A probability map is created for the grid using a label propagation algorithm based on the diffusion process and starting from a point that corresponds to these few $(x,y)$. Uncertainty is expected to increase near the phase boundary. Based on the probability values obtained at each point, a phase diagram can be created by sequentially selecting measurement points with three types of acquisition functions based on uncertainty sampling. \citet{Terayama2019} conducted an evaluation experiment where they created a three-phase diagrams of water and of Si–Al–Mg glass–ceramic glaze. They showed that the same phase diagram could be obtained with approximately 20\% of the measurements compared to using a conventional dense grid.

\subsection{Improving the efficiency of X-ray magnetic circular dichroism spectroscopy}

As an example of the improvement in measurement efficiency achieved by active learning, its application to X-ray magnetic circular dichroism (XMCD) spectroscopy by \cite{Ueno2018} is briefly introduced here. For XMCD spectroscopy, because the integrated intensity of the spectrum directly corresponds to the magnetic moment according to the magneto-optical sum rule, this can be used to analyze the magnetic properties of a material quantitatively. A precise spectrum can be obtained by fine grid of the energy measurement point, and high-precision quantitative analysis is possible by integrating the spectrum. In the past, the energy was measured at several hundred points based on the intuition and experience of the experimenter. The spectrum measurement can be formulated as a problem of estimating a function or curve $f(x)$ under the assumption that the spectral curves with the X-ray energy on the x-axis and absorption on the y-axis are connected by the function $y=f(x)+\ve$, where $\ve$ is the measurement noise. Thus, approximating the function $f(x)$ from a small number of measured points is simply a one-dimensional regression problem with active learning. \citet{Ueno2018} approximated the relationship between the X-ray energy ($x$) and absorption amount ($y$) through Gaussian process regression, and they used the predictive variance at an unobserved point of the Gaussian process regression as the acquisition function. The spectral data corresponding to 220 points measured previously were regarded as ground truth. The initial data were set to 30 randomly measured points, and a Gaussian process model was applied. Points with a large predictive variance were sequentially measured from the 220 candidate points (corresponding to pooled data) excluding the measured points. The integrated value of the Gaussian process regression curve according to the 20 new points added to the initial 30 points was confirmed to be approximately the same as the integrated value of the ground truth spectrum. In other words, the physical quantity could be measured with approximately 22\% of the number of measurements for the conventional method, which demonstrates the usefulness of active learning for material measurement. However, several issues remain to be addressed, such as the introduction of prior knowledge, dealing with measurement noise (e.g., whether or not short-term measurements should be performed multiple times), and stopping criteria.

\section{Future works and prospects}
Active learning has long history, originated in sequential experimental design, but it is still being actively researched because of its usefulness and necessity. Depending on the samples used to train the predictor, some samples may not help improve the prediction performance or even degrade it, such as in the case of incorrect labels or outliers. Although not stated in this paper, active learning has been applied to screening such samples~\citep{Abe2006}. In addition, methods such as curriculum learning~\citep{Bengio2009} and self-paced learning~\citep{Kumar2010}, where samples are selected in an appropriate order for sequential and iterative learning, can be considered as an extension of pool-based active learning.

For active learning, a small number of pre-collected samples are assumed to be available at hand. However, if the pre-collected samples are chosen according to a certain collection policy such as treatment of a particular patient, then the initial learning uses data that differ from the population distribution of the pooled data. Thus, the variance of the estimated loss function will be excessively large if the initial collected data are not handled properly. Active learning using initial data collected according to such a policy~\citep{Yan2018,Yan2019} or using a decision-making model~\citep{Sundin2019} has attracted attention in recent years. Numerous studies have used active learning to select the optimal intervention for identification in causal reasoning~\citep{Tong2001,He2008,Masegosa2013,Hauser2014}  and analyze the number of interventions (i.e., label complexity) required to identify causal relationships~\citep{Shanmugam,Greenewald2019}.

In deep learning, over-parameterization refers to the learning error tending to zero while the test error tends to decrease continuously. In this situation, bounds using learning errors are meaningless. The application of non-parametric methods (e.g., the kernel method or deep learning) may be a promising approach for reducing the problem of model bias inherent in active learning, which assumes a conventional simple hypotheses class. However, few studies have considered this approach, except for heuristic ones. For example, \citet{Karzand2020} focused on the smoothness of the function characterized by the RKHS norm of the function $h \in \cH$ to be approximated, and they proposed the acquisition function $x^{\ast} = \argmax_{x \in \cX} \min \{ \|h_{-}^{x}\|, \|h_{+}^{x}\|\}$, where $\|h_{\pm}^{x}\|$ is the RKHS norm of the hypothesis $h \in \cH$ that is updated when the label $y$ of the point $x$ is $+1$ or $-1$. They showed a case in which a label complexity of $O(\log n)$ can be achieved with this acquisition function, albeit for a special function class.
As described above, several active learning methods assume that some labeled data are available in advance, but there has been not enough discussion about the quality of the initially obtained data. Identifying a sample set for labeling from unlabeled pooled data is important for the initialization of active learning and has been discussed by \cite{Murata2020}.

\section*{Acknowledgement}
The author is supported by JST JPMJCR1761 and JPMJMI19G1. The author thank Dr. Hideaki Ishibashi, Tetsuro Ueno, Kanta Ono and Mr. Masanari Kimura for their valuable comments on the draft of this paper.

\small{

}
\label{finalpage}

\end{document}